\documentclass[runningheads]{llncs}

\usepackage{eccv}

\usepackage{eccvabbrv}

\usepackage{graphicx}
\usepackage{booktabs}
\usepackage{amssymb}
\usepackage{amsmath}
\usepackage{multirow}
\usepackage{adjustbox}
\usepackage[table]{xcolor}
\usepackage{pifont}
\usepackage{makecell}
\usepackage{subcaption}
\usepackage{colortbl}

\usepackage[accsupp]{axessibility}  

\newcommand{\cmark}{\ding{51}}
\newcommand{\xmark}{\ding{55}}
\DeclareMathOperator*{\argmax}{argmax}

\definecolor{table1}{HTML}{ffe5ec}
\definecolor{table2}{HTML}{FFF9C0}
\definecolor{table3}{HTML}{D9E9FF}
\definecolor{table4}{HTML}{DFFFF0}
\definecolor{table5}{HTML}{E1EDF6}
\definecolor{table6}{HTML}{caf0f8}

\usepackage{hyperref}

\usepackage{orcidlink}
\begin{document}

\title{SVMemAgent: A Streaming Video Memory Agent for Query-Agnostic Online Frame Selection} 

\titlerunning{SVMemAgent}

\author{
Dohwan Ko\inst{1, 2}$^\dagger$ \and
Ji Soo Lee\inst{3} \and
Pierce Chuang\inst{2} \and
Debojeet Chatterjee\inst{2} \and
Ashish Shenoy\inst{2} \and
Yichao Lu\inst{2} \and
Seungwhan Moon\inst{2} \and
Xin Luna Dong\inst{2} \and
Vikas Bhardwaj\inst{2} \and
Hyunwoo J. Kim\inst{3}$^*$
}

\authorrunning{D. Ko et al.}

\institute{
Korea University, Seoul, Republic of Korea \and
Meta AI \and
KAIST, Daejeon, Republic of Korea
}

\maketitle

{\let\thefootnote\relax\footnotetext{$^\dagger$ Work done at Meta. \; $^*$ Corresponding author.}}

\begin{abstract}
    Most keyframe selection studies focus on offline settings, assuming access to the full video and query in advance. 
    In contrast, real-world streaming scenarios require online frame selection under unknown video duration, without access to either the query or future frames during selection.
    To address this, we introduce Streaming Video Memory (SVMem), a compact and representative memory of previously observed content, updated continuously as the video stream unfolds. 
    Building on this setting, we propose the Streaming Video Memory Agent (SVMemAgent), which dynamically maintains a memory by deciding at each timestep whether to replace an existing memory frame with the incoming frame or discard it.
    SVMemAgent is trained using Group Relative Policy Optimization (GRPO) with task-driven rewards derived from diverse question-answer pairs, implicitly exposing the policy to a distribution of queries during training so that SVMem retains generally informative frames at inference, when queries are unavailable.    
    Experiments on both online and offline video benchmarks show that SVMemAgent consistently outperforms online frame selection baselines and achieves competitive performance with offline methods that assume access to the full video and query.
    Through task-driven rewards, SVMemAgent learns an emergent keyframe selection policy that prefers frames containing textual information, which may benefit downstream VideoQA tasks.
    \keywords{Online Video Understanding \and Keyframe Selection \and Memory Agent \and Group Relative Policy Optimization}
\end{abstract}
\section{Introduction}

\begin{figure}[!t] 
    \centering
    \begin{subfigure}[b]{0.46\linewidth}
        \centering
        \includegraphics[width=\linewidth]{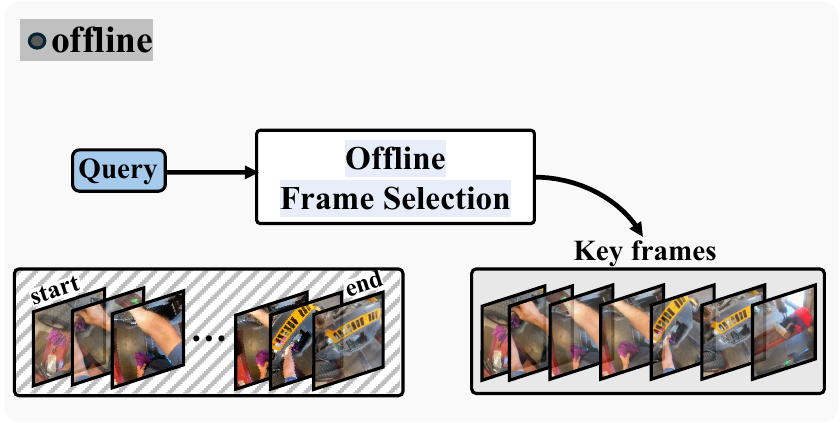}
        \caption{\textbf{Offline frame selection methods.}}
        \label{fig:intro-1}
    \end{subfigure}
    \begin{subfigure}[b]{0.53\linewidth}
        \centering
        \includegraphics[width=\linewidth]{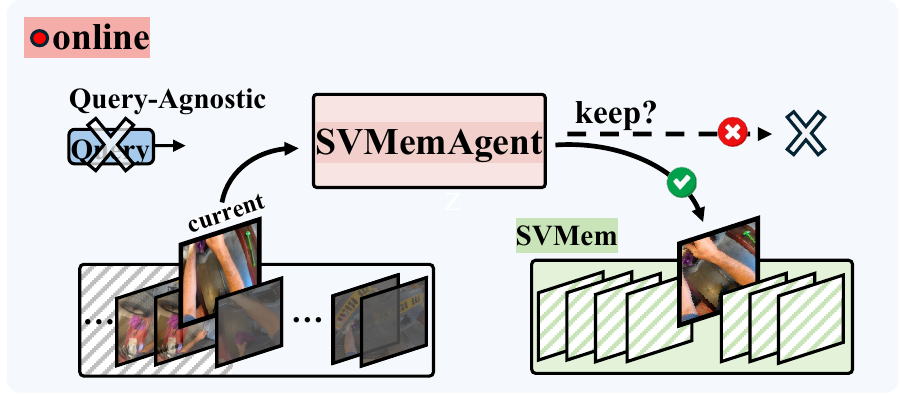}
        \caption{\textbf{Online frame selection with SVMemAgent.}}
        \label{fig:intro-2}
    \end{subfigure}
    \caption{
        (a) Offline frame selection methods simply retrieve query-relevant frames with access to the full video, whereas (b) SVMemAgent sequentially updates SVMem without access to the query or future frames, deciding whether to discard the current frame or use it to replace an existing memory slot.
    }
    \label{fig:intro}
\end{figure}
Keyframe selection in long video understanding enables Video Large Language Models (VideoLLMs)~\cite{li2024videochat,li2025videochat,ko2023large,zhang2024video,zhang2025videollama} to process long videos within limited context windows. 
Existing approaches mainly study this problem in offline settings, where the complete video and the corresponding query are available in advance. 
Under this assumption, frame selection can be naturally formulated as a simple retrieval problem, which selects query-relevant frames from the full video (Fig.~\ref{fig:intro-1}). 
However, this offline setting overlooks real-world streaming scenarios, such as wearable devices and embodied AI systems, where frames arrive sequentially, the video duration is unknown, and the query may be unavailable at the time of frame selection. 
As a result, in streaming settings, frames must be selected sequentially without access to future visual content or the user query.

Sequential frame selection in online streaming scenarios poses two key challenges. 
First, since the total stream duration is unknown during streaming, naive fixed-interval frame sampling cannot guarantee that the memory will remain below capacity before the query arrives.
Second, as frame selection occurs prior to query arrival, decisions must rely solely on past and incoming frames, without access to future frames or query information.
These constraints make existing offline frame selection approaches, which merely identify query-relevant frames from the entire video, unsuitable for direct application to online settings.

To address these challenges, we first formulate a Streaming Video Memory (\textbf{SVMem}), which maintains a subset of frames without storing the entire video stream. 
As in Fig.~\ref{fig:intro-2}, SVMem is sequentially updated by \textbf{SVMemAgent}, which decides whether to replace an existing memory frame with the incoming frame or discard it, without access to query information or future frames.
Once streaming ends and a query is provided, the downstream VideoLLM processes the query using the frames stored in the final state of SVMem.
We train SVMemAgent using Group Relative Policy Optimization (GRPO)~\cite{shao2024deepseekmath} with task-driven rewards. 
This enables SVMemAgent to learn a general prior over informative frames from the training query distribution through reward signals, despite having no access to the inference-time query during frame selection.
Furthermore, we introduce Diversity-Aware Advantage Discounting (DAAD), which penalizes policies exhibiting low reward variance despite high state diversity.
This mitigates overfitting to uninformative reward signals that are largely insensitive to frame-selection decisions, such as language-biased queries or static scenes.
Under the strict online streaming setting, SVMemAgent consistently outperforms online frame selection baselines and even achieves competitive performance with offline methods that assume access to the full video and query.
We observe that SVMemAgent implicitly learns to prioritize informative and non-redundant frames, such as text-containing frames, for future query answering.

To sum up, our contributions are \textbf{threefold}:
(1) We introduce SVMem, a memory for a strict online streaming setting that requires frame selection under unknown duration, limited memory capacity, and no access to query information.
(2) We propose SVMemAgent, which maintains a compact and representative memory by discarding or replacing frames as the stream unfolds. 
We train the agent using GRPO with task-driven rewards and DAAD regularization.
(3) Extensive experiments on both online and offline video benchmarks demonstrate that SVMemAgent consistently outperforms existing frame selection baselines and achieves performance competitive with offline methods that assume access to the full video and query.
\section{Related Works}

\noindent \textbf{Online video understanding.}
Recent studies have begun applying VideoLLMs to online streaming settings.
One line of work~\cite{chen2024videollm,huang2024online,yao2025timechat,fu2025vispeak,qian2025dispider} adopts a data-centric approach, introducing instruction-following datasets to train VideoLLMs for generating timely responses to streaming videos. 
For example, VideoLLM-online~\cite{chen2024videollm} proposes a narration streaming dataset that encourages proactive description generation based on sequences of user actions.
Another line of work~\cite{di2025streaming,qian2024streaming,li2025lion,liu2024streamchat,xiong2025streaming} focuses on architectural modifications to enable streaming video processing. 
StreamChat~\cite{xiong2025streaming}, for instance, incorporates both long- and short-term memory modules to efficiently handle queries in long videos. 
However, most existing approaches rely on training-free heuristics for memory maintenance, whereas our memory agent directly learns a query-agnostic frame selection policy through reinforcement learning.

\noindent \textbf{Keyframe selection in video understanding.}
Keyframe selection identifies a compact subset of informative frames, enabling VideoLLMs to process long videos under limited context windows. 
Existing approaches can be categorized according to the information available at selection time. 
Offline methods~\cite{liu2024streamchat,tang2025adaptive,hu2025m} assume access to either the full video or the user query. 
For example, AKS~\cite{tang2025adaptive} recursively partitions the video into a hierarchical structure and selects keyframes based on CLIP-based query-frame relevance. 
In online streaming settings~\cite{yao2025timechat,he2024ma,zhang2025flash}, however, frames must be selected without prior knowledge of either the query or the total video duration. 
TimeChat-Online~\cite{yao2025timechat} removes redundant frames by detecting feature changes relative to the most recently retained frame, while MA-LMM~\cite{he2024ma} iteratively consolidates adjacent frame pairs within a fixed-capacity memory bank. 

\noindent \textbf{Reinforcement learning for visual reasoning.}
Inspired by the success of reinforcement learning (RL)-based post-training for LLMs, particularly Group Relative Policy Optimization (GRPO)~\cite{shao2024deepseekmath}, recent studies have extended RL with verifiable rewards to multi-modal models~\cite{liu2025visual,feng2026video,li2025videochat,feng2026onethinker}. 
Visual-RFT~\cite{liu2025visual} applies GRPO with task-specific verifiable rewards to visual perception tasks, including detection and grounding, and demonstrates substantially greater data efficiency than supervised fine-tuning. 
In the video domain, Video-R1~\cite{feng2026video} introduces a temporally aware variant of GRPO to incentivize temporal reasoning in VideoLLMs. 
However, these approaches primarily use RL to enhance the reasoning capability of the VideoLLM itself and assume full access to the input video, while we employ GRPO to train a lightweight external policy for online memory maintenance.
Task-driven rewards derived from diverse question-answer pairs expose the policy to a broad query distribution during training, enabling it to learn a query-agnostic prior for retaining informative frames under strict online streaming constraints.
\section{Method}

SVMemAgent sequentially updates SVMem during video streaming.
It first decides to discard or retain the current frame and, if retained, determines which memory slot to replace.
We first describe the overall design of SVMem and SVMemAgent, followed by detailed training and inference pipelines.

\begin{figure*}[!t] 
\centering
\includegraphics[width=\linewidth]{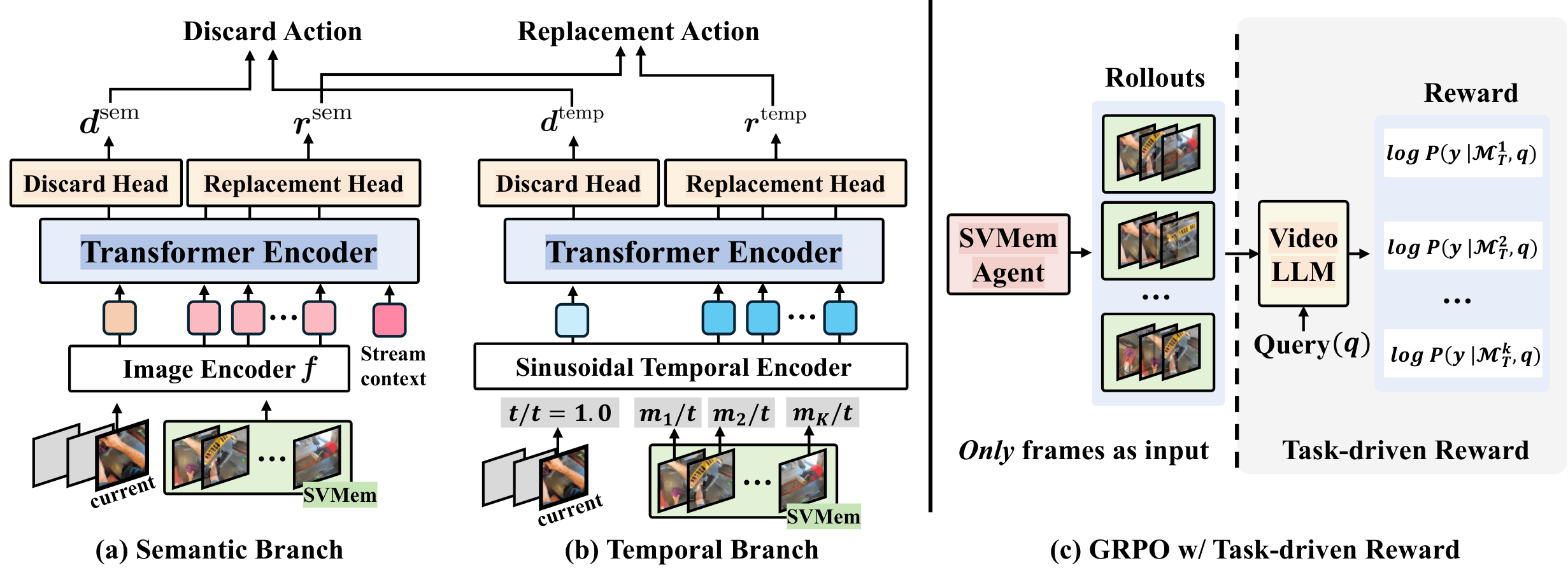}
    \caption{
        \textbf{Illustration of SVMemAgent and GRPO training with task-driven rewards.}
        SVMemAgent integrates (a) a semantic branch designed for content-aware memory updates and (b) a temporal branch for maintaining uniform temporal coverage.
        (c) During training, SVMemAgent is optimized using GRPO with task-driven rewards to learn a general prior over informative frames without requiring query information at inference time.
    }
    \label{fig:main}
\end{figure*}
\subsection{SVMem}

We consider a strict online streaming video understanding setting under two constraints: (1) video frames arrive sequentially at 1 FPS, while future frames and the total video duration remain inaccessible, and (2) the user query is unavailable during frame selection, resulting in a query-agnostic setting. 
Under these constraints, we introduce an SVMem $\mathcal{M}$ that stores at most $K$ frames.
At each timestep $t$, a new frame $v_t$ arrives, and the system must decide whether to discard $v_t$ or replace one of the existing $K$ memory slots with it. 
When the query arrives at an arbitrary timestamp, the corresponding memory state $\mathcal{M} = \{v_{m_1}, v_{m_2}, \ldots, v_{m_K}\}$ and the user query are provided to a VideoLLM, such as InternVL3.5~\cite{wang2025internvl3} or Qwen3-VL~\cite{bai2025qwen3}, for answer generation.
\subsection{SVMemAgent}

To update SVMem sequentially during streaming, we introduce SVMemAgent.
Since neither the query nor the total stream duration is available during memory updates, SVMemAgent jointly considers both \textit{what} a frame depicts and \textit{when} it occurs within the observed stream. 
Accordingly, SVMemAgent adopts a dual-branch architecture that models \emph{semantic} understanding and \emph{temporal} coverage, as in Fig.~\ref{fig:main}a and \ref{fig:main}b.
It adopts a two-stage action policy that first makes a \emph{discard} decision and then, if necessary, performs a \emph{replacement} decision.

\noindent \textbf{Dual-branch decision module.}
In the \emph{semantic branch}, the image encoder $f$ first extracts a visual feature $e_t = f(v_t) \in \mathbb{R}^{D}$ from each incoming frame $v_t$, where $D$ denotes the hidden dimension. 
Lightweight transformer layers then model semantic relationships among the current frame, the existing memory slots, and the global stream context.
Specifically, the transformer processes $K+2$ input tokens: (1) \textbf{Frames in SVMem} ($K$ tokens)-visual features of frames currently stored in memory, $\{e_{m_k}\}_{k=1}^{K}$, (2) \textbf{Current frame} (1 token)-feature of the incoming frame, $e_t$, and (3) \
\textbf{Stream context} (1 token)-average feature of all previously observed frames, which serves as a summary of the video stream, $c_t = \frac{1}{t} \sum_{i=0}^{t-1} e_i$.
Two prediction heads are applied to the transformer outputs to produce discard and replacement scores. 
The discard head uses the representation corresponding to the current frame token to generate a scalar discard score $d^{\text{sem}} \in \mathbb{R}$. 
The replacement head uses the representations of the $K$ memory slot tokens to produce replacement scores $r^{\text{sem}} \in \mathbb{R}^{K}$.

The \emph{temporal branch} uses only frame timestamps, without relying on semantic features, to encourage temporal coverage of the observed stream up to the current timestamp.
It processes $K+1$ input tokens corresponding to the $K$ memory frames and the current frame.
Specifically, each memory slot is represented by its normalized timestamp $m_k / t \in [0,1)$, while the current frame is assigned the fixed position $t / t = 1.0$.
These scalar timestamps are transformed into $D$-dimensional embeddings using sinusoidal positional encodings followed by a linear projection layer. 
The resulting tokens are processed by shallow transformer layers.
Discard and replacement heads are then applied to produce temporal discard and replacement scores, denoted by $d^{\text{temp}}$ and $r^{\text{temp}}$, respectively.

The dual-branch outputs are combined as:
\begin{align}
    d & = w_d \cdot d^{\text{sem}} + (1 - w_d) \cdot d^{\text{temp}} \in \mathbb{R}, \\
    r & = w_r \cdot r^{\text{sem}} + (1 - w_r) \cdot r^{\text{temp}} \in \mathbb{R}^K,
\end{align}
where $w_d, w_r \in [0,1]$ denote the balancing weights for discard and replacement decisions, respectively.

\noindent \textbf{Two-stage action decision.}
At each timestep, SVMemAgent first decides the discard action based on $P_d = \text{sigmoid}(d)$, \ie, discard action = \texttt{`discard'} if $P_d > 0.5$ else \texttt{`retain'}.
When the discard action is \texttt{`retain'}, the memory slot with the highest replacement probability under $P_r(k) = \text{softmax}(r)_k$ is selected and replaced by the current frame, \ie, replacement action = $\argmax_k P_r(k)$.
During GRPO training (Sec.~\ref{subsec:training}), both actions are instead sampled stochastically from $P_d$ and $P_r$ to enable policy exploration.
\subsection{Training Pipeline}
\label{subsec:training}

SVMemAgent is trained in two stages: (1) cold-start training with pseudo-labels to initialize the dual-branch policy, and (2) GRPO, which refines the policy using task-driven rewards with Diversity-Aware Advantage Discounting (DAAD).

\noindent \textbf{Stage 1: Cold-start training with pseudo-labels.}
Since ground-truth annotations for query-agnostic online frame selection are unavailable, we construct heuristic pseudo-labels separately for each decision branch. 
The pseudo-labels for the \emph{semantic branch} are designed to preserve frames that are representative of the video stream while remaining non-redundant with respect to the current memory contents.
To this end, we adopt a teacher visual encoder $g$ to compute relevance and novelty scores based on cosine similarity. 
The current frame is retained if it is sufficiently representative of the video stream (relevance) and contributes information not already captured in memory (novelty). 
Specifically, the discard pseudo-label is assigned as 0 (retain) when the sum of the relevance and novelty scores exceeds a predefined threshold, and as 1 (discard) otherwise. 
If the current frame is retained (\ie, not discarded), the replacement pseudo-label is assigned to the index of the memory slot most similar to the current frame, thereby reducing redundancy in memory.

While the semantic branch pseudo-labels focus on preserving informative and non-redundant visual content, the \emph{temporal branch} pseudo-labels promote balanced temporal coverage throughout the video stream.
At each timestep $t$, the discard pseudo-label is assigned when the memory already provides sufficiently uniform temporal coverage.
Otherwise, we evaluate all possible replacement actions and assign the replacement pseudo-label to the candidate that maximizes temporal uniformity.
Each branch is trained individually using its corresponding pseudo-labels.
The discard head uses a BCE loss for discard/retain decisions, while the replacement head is trained with a CE loss over the $K$ memory slots.

\noindent \textbf{Stage 2: GRPO with DAAD.}
While the cold-start training provides a stable policy initialization, pseudo-labels do not directly optimize downstream VideoQA performance.
Therefore, we further optimize SVMemAgent using GRPO with task-driven rewards, as in Fig.~\ref{fig:main}c. 
Since SVMemAgent has no access to the query during frame selection at inference time, we train it to implicitly leverage the training query distribution through the task-driven reward signals.
This enables the agent to retain frames that are broadly useful across diverse potential queries and thereby learn a query-agnostic prior over visual importance.
In GRPO, actions are sampled from both the discard and replacement heads, and the corresponding policies are jointly optimized.

To evaluate the quality of frame selection decisions, we define task-driven rewards by measuring how effectively the selected frames support answering a user query. 
Specifically, for the $i$-th rollout, we feed the final memory state $\mathcal{M}_T^i$, containing $K$ frames, and the query $q$, into a frozen VideoLLM reward model. 
The reward is defined as the average per-token log-probability assigned to the ground-truth response $y$:
\begin{equation}
    R^i = \frac{1}{|y|} \sum_{j=1}^{|y|} \log P_{\text{VideoLLM}}\left(y_j \mid y_{<j}, \mathcal{M}_T^i, q\right),
\end{equation}
which corresponds to the negative of the average token-level cross-entropy loss.
This objective directly aligns policy optimization with downstream VideoQA performance.
Since SVMemAgent is designed for a query-agnostic streaming setting, optimizing against a single query-answer pair may lead to overfitting to query-specific visual cues. 
Therefore, we use four distinct queries per video to encourage broader generalization.

However, we observe that substantially diverse memory states can lead to nearly identical rewards, indicating that the reward signal is largely insensitive to frame selection decisions. 
This phenomenon can arise when a query is answerable from language priors alone (language-biased query) or when distinct frame subsets contain semantically equivalent information, particularly in static scenes.
Regardless of the underlying cause, such rollout groups provide weak supervision for frame selection and inject noise into policy optimization, thereby hindering the agent from learning a meaningful selection policy.

To address this issue, we propose \textbf{Diversity-Aware Advantage Discounting (DAAD)}.
We first compute a state diversity metric as the mean pairwise Jaccard distance between final memory states:
\begin{equation}
    D_{\text{Jac}} = \frac{2}{G(G-1)}\sum_{g < g'}\left(1 - \frac{|\mathcal{M}_T^{(g)} \cap \mathcal{M}_T^{(g')}|}{|\mathcal{M}_T^{(g)}\cup \mathcal{M}_T^{(g')}|}
\right),
\end{equation}
where $G$ is the group size. 
Then, we define an advantage discounting score $\alpha$ that quantifies the ratio of memory diversity to reward variation:
\begin{equation}
    \alpha = \text{sigmoid}\left(\lambda - \frac{D_{\text{Jac}}}{\text{std}\left(\{R^j\}_{j=1}^{G}\right)}\right),
\end{equation}
where $\lambda$ is a predefined threshold.
$\alpha$ adaptively rescales the normalized GRPO advantages:
\begin{equation}
    \hat{A}^i = \alpha \cdot \frac{R^i - \text{mean}\left(\{R^j\}_{j=1}^{G}\right)}{\text{std}\left(\{R^j\}_{j=1}^{G}\right)}.
\label{eq:aiad}
\end{equation}
When sampled policies generate highly diverse memory states but exhibit only minor reward differences, the optimization signal is likely dominated by noise rather than meaningful policy improvements. 
In such cases, $\alpha$ approaches 0, effectively suppressing unreliable policy gradients. 
Conversely, when reward variations consistently reflect differences in memory quality, $\alpha$ approaches 1, thereby preserving the original GRPO advantages.
\begin{table*}[t]
    \centering
    \small
    \setlength{\tabcolsep}{4pt}
    \caption{
        \textbf{Comparison of frame selection methods.} 
        Red rows indicate baselines evaluated under relaxed online or offline settings, where the query, the video duration, or both are known in advance. 
        Blue rows denote baselines evaluated under the strict online setting, where both the query and video duration are unknown.
    }
    \begin{adjustbox}{width=\textwidth}
    \begin{tabular}{l|c c|l}
        \toprule
        \textbf{Method} & \makecell{\textbf{Unknown}\\\textbf{Query}} & \makecell{\textbf{Unknown}\\\textbf{Duration}} & \textbf{Description} \\
        \midrule
        \rowcolor{table1}
        Uniform & \cmark & \xmark & Evenly spaced frames across the full video \\
        \rowcolor{table1}
        Random  & \cmark & \xmark & Randomly sampled frames from the full video \\
        \rowcolor{table1}
        Clustering  & \cmark & \xmark & $K$-means over the full video; frames nearest to centroids \\
        \rowcolor{table1}
        StreamChat~\cite{xiong2025streaming} & \xmark & \cmark & Ebbinghaus memory + $K$-means compression + CLIP retrieval \\
        \rowcolor{table1}
        AKS~\cite{tang2025adaptive}  & \xmark & \xmark & Query-aware recursive hierarchical keyframe splitting based on CLIP \\
        \rowcolor{table1}
        M-LLM Selector~\cite{hu2025m}  & \xmark & \xmark & A distilled CLIP model trained to predict query-frame relevance scores. \\
        \midrule
        \rowcolor{table3}
        FIFO & \cmark & \cmark & First-in-first-out memory update with fixed-interval sampling \\
        \rowcolor{table3}
        Reservoir & \cmark & \cmark & Vitter's Algorithm R with decreasing replace probability \\
        \rowcolor{table3}
        TimeChat-Online~\cite{yao2025timechat} & \cmark & \cmark & Feature change detection against last kept frame \\
        \rowcolor{table3}
        Flash-VStream~\cite{zhang2025flash} & \cmark & \cmark & PCA + temporally-ordered sequential $K$-means clustering \\
        \rowcolor{table3}
        MA-LMM~\cite{he2024ma} & \cmark & \cmark & Adjacent-pair consolidation in fixed memory bank \\
        \midrule
        \rowcolor{table3}
        \textbf{SVMemAgent (Ours)}  & \cmark & \cmark & A streaming video memory agent with sequential discard-and-replacement policy decisions \\
        \bottomrule
    \end{tabular}
    \end{adjustbox}
    \label{tab:baselines}
\end{table*}
\subsection{Inference Pipeline}

Unlike training, which relies on stochastic action sampling for policy exploration, inference uses a deterministic greedy policy to construct the memory state. 
Specifically, the discard decision is computed as $\mathbf{1}_{[P_d > 0.5]}$, while the replacement slot is selected as $\argmax_k P_r(k)$. 
When the query arrives at an arbitrary timestamp, the corresponding memory state and query are passed to the downstream VideoLLM to generate the response.
We first store the initial $K$ frames to fill the memory and activate SVMemAgent once the memory reaches full capacity. 
If a query arrives before the memory is full, we use all frames observed up to that timestamp.
\section{Experiments}

We evaluate our method on both offline and online video understanding benchmarks. 
Offline benchmarks pose the query after the stream ends, whereas online benchmarks pose queries at arbitrary timestamps during streaming.
During evaluation, SVMemAgent processes videos strictly in an online manner at 1 FPS, without access to future frames, the total video duration, or the user query during frame selection. 
The memory state available when the query arrives is provided to the downstream VideoLLMs, InternVL3.5-8B~\cite{wang2025internvl3} and Qwen3-VL-8B~\cite{bai2025qwen3}.

\noindent \textbf{Implementation details.}
In SVMemAgent, we adopt DINOv3-Base~\cite{simeoni2025dinov3} as the image encoder $f$, with 8-layer and 2-layer transformer encoders for the semantic and temporal branches, respectively, resulting in a total of 196M parameters.
During training, the maximum memory budget of SVMem is set to $K = 16$.
We use subsets of video-question-answer samples from the Video-R1~\cite{feng2026video} and LongVILA~\cite{chen2024longvila} datasets.
We filter out videos shorter than 64 seconds.
In cold-start training, DINOv3-Huge~\cite{simeoni2025dinov3} is employed as the teacher image encoder $g$ for pseudo-label generation.
For GRPO training, we use a rollout group size $G$ of 32 and set the DAAD threshold to $\lambda = 40$. 
We use InternVL3.5-2B~\cite{wang2025internvl3} as the VideoLLM reward model for task-driven supervision.

\noindent \textbf{Baselines.}
We compare SVMemAgent against a broad range of heuristic and state-of-the-art frame-selection baselines.
For a fair comparison, we categorize these baselines according to whether they satisfy the core constraints of our problem, \ie, query-agnostic frame selection in an online streaming setting with unknown video duration.
Directly comparable baselines are evaluated under the same strict online setting and memory budget, without access to future frames or the user query. 
In contrast, offline and query-aware methods are reported separately.
We summarize all baselines in Tab.~\ref{tab:baselines}.

\begin{table*}[t]
    \centering
    \caption{\textbf{Comparison with baselines of the strict online streaming setting.}}
    \setlength{\tabcolsep}{3pt}
    \begin{adjustbox}{width=0.99\linewidth}
    \begin{tabular}{l l|c c c|c|c}
        \toprule
        \multirow{2}{*}{\textbf{Model}}
        & \multirow{2}{*}{\textbf{Methods}}
        & \multicolumn{3}{c|}{\textbf{Offline video benchmarks}}
        & \multicolumn{1}{c|}{\textbf{Online video benchmark}}
        & \multirow{2}{*}{\textbf{Average}} \\
        \cmidrule(lr){3-5} \cmidrule(lr){6-6}
        & & VideoMME & LongVideoBench & EgoSchema
        & RVS-Ego & \\
        \midrule
        \midrule
        \multirow{6}{*}{InternVL3.5-8B}
        & FIFO & 55.8 & 55.9 & 51.6 & 51.8 & \cellcolor{table6}{53.8} \\
        & Reservoir & 60.4 & 57.7 & 54.8 & 53.1 & \cellcolor{table6}{56.5} \\
        & TimeChat-Online~\cite{yao2025timechat} & 59.7 & 58.0 & 53.8 & 49.0 & \cellcolor{table6}{55.1} \\
        & Flash-VStream~\cite{zhang2025flash} & 57.3 & 57.9 & 54.2 & 53.5 & \cellcolor{table6}{55.7} \\
        & MA-LMM~\cite{he2024ma} & 59.5 & 57.5 & 54.4 & 51.0 & \cellcolor{table6}{55.6} \\
        \cmidrule{2-7}
        \rowcolor{table5}
        & \textbf{SVMemAgent (ours)} & \textbf{61.5} & \textbf{58.5} & \textbf{55.6} & \textbf{54.1} & \cellcolor{table6}\textbf{57.4} \\
        \midrule
        \midrule
        \multirow{6}{*}{Qwen3-VL-8B}
        & FIFO & 54.4 & 55.6 & 66.7 & 57.3 & \cellcolor{table6}{58.5} \\
        & Reservoir & 59.1 & 57.8 & 73.6 & 58.2 & \cellcolor{table6}{62.2} \\
        & TimeChat-Online~\cite{yao2025timechat} & 59.3 & 57.7 & 72.0 & 55.4 & \cellcolor{table6}{61.1} \\
        & Flash-VStream~\cite{zhang2025flash} & 57.2 & 57.0 & 72.0 & \textbf{58.8} & \cellcolor{table6}{61.2} \\
        & MA-LMM~\cite{he2024ma} & 59.2 & 57.1 & 72.4 & 58.6 & \cellcolor{table6}{61.8} \\
        \cmidrule{2-7}
        \rowcolor{table5}
        & \textbf{SVMemAgent (ours)} & \textbf{60.6} & \textbf{58.0} & \textbf{76.2} & \textbf{58.8} & \cellcolor{table6}\textbf{63.4} \\
        \bottomrule
    \end{tabular}
    \end{adjustbox}
    \label{tab:main}
\end{table*}
\subsection{Results on Video Benchmarks}

\noindent \textbf{Comparison with strict online streaming baselines.}
Tab.~\ref{tab:main} compares SVMemAgent with both heuristic and state-of-the-art frame selection baselines under the strict online streaming setting, where future frames, total video duration, and query information are unavailable during frame selection. 
Under these constraints, SVMemAgent consistently outperforms competing approaches in terms of average accuracy.
For example, when integrated with InternVL3.5-8B, SVMemAgent improves the average accuracy over Flash-VStream by 1.7\%. 
Notably, although SVMemAgent is trained using InternVL3.5-2B as the reward model during GRPO, it generalizes effectively to a different backbone VideoLLM, Qwen3-VL-8B. 
In particular, SVMemAgent surpasses TimeChat-Online by 2.3\% in average accuracy, demonstrating cross-model transferability of the learned memory policy.

\noindent \textbf{Comparison with relaxed online or offline baselines.}
In Tab.~\ref{tab:offline}, we further compare SVMemAgent with baselines evaluated under relaxed online or offline settings, where additional information, such as the total video duration, query information, or both, is available during frame selection. 
Despite having access to substantially less information, SVMemAgent outperforms most competing approaches. 
In particular, SVMemAgent outperforms StreamChat, which uses query information for keyframe selection, by 1.8\% when evaluated with InternVL3.5-8B, despite having no access to the user query during frame selection.
Overall, these results suggest that SVMemAgent learns an adaptive memory maintenance policy that sequentially discards and replaces frames under strict online constraints while achieving performance comparable to methods with substantially stronger information access.
\begin{table*}[t]
    \centering
    \caption{\textbf{Comparison with baselines under relaxed online or offline settings.}}
    \setlength{\tabcolsep}{3pt}
    \begin{adjustbox}{width=0.99\linewidth}
    \begin{tabular}{l l|c c|c c c|c|c}
        \toprule
        \multirow{4}{*}{\textbf{Model}}
        & \multirow{4}{*}{\textbf{Methods}}
        & \multicolumn{2}{c|}{\textbf{Setting assumptions}}
        & \multicolumn{3}{c|}{\textbf{Offline video benchmarks}}
        & \multicolumn{1}{c|}{\textbf{Online video benchmark}}
        & \multirow{4}{*}{\textbf{Average}} \\
        \cmidrule(lr){3-4}
        \cmidrule(lr){5-7}
        \cmidrule(lr){8-8}
        &
        & \makecell{\textbf{Unknown}\\\textbf{Query}}
        & \makecell{\textbf{Unknown}\\\textbf{Duration}}
        & \textbf{VideoMME}
        & \textbf{LongVideoBench}
        & \textbf{EgoSchema}
        & \textbf{RVS-Ego}
        & \\
        \midrule
        \midrule

        \multirow{7}{*}{InternVL3.5-8B}
        & Uniform
        & \cmark & \xmark
        & 61.3 & 57.4 & 55.4 & 53.9
        & \cellcolor{table6}{57.0} \\

        & Random
        & \cmark & \xmark
        & 59.7 & 57.5 & 55.4 & 53.9
        & \cellcolor{table6}{56.6} \\

        & Clustering
        & \cmark & \xmark
        & \textbf{62.5} & 58.4 & 55.0 & 53.0
        & \cellcolor{table6}{57.2} \\

        & StreamChat~\cite{xiong2025streaming}
        & \xmark & \cmark
        & 58.9 & 59.3 & 52.2 & 52.0
        & \cellcolor{table6}{55.6} \\

        & AKS~\cite{tang2025adaptive}
        & \xmark & \xmark
        & 61.6 & \textbf{60.4} & 54.4 & 53.5
        & \cellcolor{table6}{\textbf{57.5}} \\

        & M-LLM Selector~\cite{hu2025m}
        & \xmark & \xmark
        & 53.8 & 51.9 & 49.0 & 42.3
        & \cellcolor{table6}{49.2} \\

        \cmidrule{2-9}
        \rowcolor{table5}
        & \textbf{SVMemAgent (ours)}
        & \cmark & \cmark
        & 61.5 & 58.5 & \textbf{55.6} & \textbf{54.1}
        & \cellcolor{table6}{57.4} \\

        \midrule
        \midrule

        \multirow{7}{*}{Qwen3-VL-8B}
        & Uniform
        & \cmark & \xmark
        & 61.1 & 58.3 & 75.2 & 58.5
        & \cellcolor{table6}{63.3} \\

        & Random
        & \cmark & \xmark
        & 59.8 & 57.1 & 72.2 & 58.8
        & \cellcolor{table6}{62.0} \\

        & Clustering
        & \cmark & \xmark
        & \textbf{62.2} & 58.0 & 75.0 & \textbf{59.7}
        & \cellcolor{table6}{63.7} \\

        & StreamChat~\cite{xiong2025streaming}
        & \xmark & \cmark
        & 59.4 & 60.3 & 71.8 & 58.1
        & \cellcolor{table6}{62.4} \\

        & AKS~\cite{tang2025adaptive}
        & \xmark & \xmark
        & 61.6 & \textbf{61.1} & 75.0 & 59.0
        & \cellcolor{table6}{\textbf{64.2}} \\

        & M-LLM Selector~\cite{hu2025m}
        & \xmark & \xmark
        & 52.0 & 50.4 & 62.5 & 41.0
        & \cellcolor{table6}{51.5} \\

        \cmidrule{2-9}
        \rowcolor{table5}
        & \textbf{SVMemAgent (ours)}
        & \cmark & \cmark
        & 60.6 & 58.0 & \textbf{76.2} & 58.8
        & \cellcolor{table6}{63.4} \\

        \bottomrule
    \end{tabular}
    \end{adjustbox}
    \label{tab:offline}
\end{table*}
\subsection{Ablation Studies}

We conduct ablation studies on each training stage of SVMemAgent and analyze its temporal and semantic branches. 
We further evaluate inference-time performance under varying memory budgets.

\begin{table*}[!t]
    \centering
    \begin{minipage}[t]{0.55\textwidth}
        \centering
        \captionof{table}{\textbf{Effect of each training stage.}}
        \begin{adjustbox}{width=\linewidth}
        \begin{tabular}{l|c c|c c|c}
            \toprule
            \textbf{Model} & \textbf{Cold-start} & \textbf{GRPO} & \textbf{Offline} & \textbf{Online} & \textbf{Average} \\
            \midrule
            \midrule
            \multirow{2}{*}{InternVL3.5-8B}
            & \ding{52} &  & 57.9 & 52.9 & 56.7 \\
            & \ding{52} & \ding{52} & \textbf{58.5} & \textbf{54.1} & \textbf{57.5} \\
            \midrule
            \multirow{2}{*}{Qwen3-VL-8B}
            & \ding{52} &  & 64.1 & 58.2 & 62.6 \\
            & \ding{52} & \ding{52} & \textbf{64.9} & \textbf{58.8} & \textbf{63.4} \\
            \bottomrule
        \end{tabular}
        \end{adjustbox}
        \label{tab:stage}
    \end{minipage}
    \hfill
    \begin{minipage}[t]{0.43\textwidth}
        \centering
        \captionof{table}{\textbf{Effect of each branch.}}
        \begin{adjustbox}{width=\linewidth}
        \begin{tabular}{l|c c|c c|c}
            \toprule
            \textbf{Model} & \textbf{Temporal} & \textbf{Semantic} & \textbf{Offline} & \textbf{Online} & \textbf{Average} \\
            \midrule
            \midrule
            \multirow{3}{*}{InternVL3.5-8B}
            & \ding{52} &  & 57.6 & 54.1 & 56.8 \\
            &  & \ding{52} & 55.5 & 51.4 & 54.5 \\
            & \ding{52} & \ding{52} & \textbf{58.5} & \textbf{54.1} & \textbf{57.5} \\
            \midrule
            \multirow{3}{*}{Qwen3-VL-8B}
            & \ding{52} &  & 63.7 & 58.2 & 62.4 \\
            &  & \ding{52} & 62.2 & 54.6 & 60.3 \\
            & \ding{52} & \ding{52} & \textbf{64.9} & \textbf{58.8} & \textbf{63.4} \\
            \bottomrule
        \end{tabular}
        \end{adjustbox}
        \label{tab:branch}
    \end{minipage}
\end{table*}

\noindent \textbf{Performance of each training stage.}
In Tab.~\ref{tab:stage}, applying GRPO after cold-start training consistently improves performance across both backbones, leading to average gains of 0.8\% . 
This suggests that task-driven reward supervision in GRPO refines the policy beyond heuristic pseudo-label supervision, enabling the agent to select frames that are broadly useful across diverse queries.

\noindent \textbf{Effect of each branch.}
As shown in Tab.~\ref{tab:branch}, the temporal branch alone outperforms the semantic branch alone, indicating that maintaining temporal uniformity in memory provides a robust strategy for query-agnostic memory maintenance when the video duration is unknown.
Combining the two branches consistently achieves the best results across both backbones, highlighting the necessity of jointly considering temporal coverage and informative content preservation.

\begin{table*}[t]
    \centering
    \caption{\textbf{Results with different reward models.}}
    \setlength{\tabcolsep}{3pt}
    \begin{adjustbox}{width=0.99\linewidth}
    \begin{tabular}{l l|c c c|c|c}
        \toprule
        \multirow{2}{*}{\textbf{Model}}
        & \multirow{2}{*}{\textbf{Reward model}}
        & \multicolumn{3}{c|}{\textbf{Offline video benchmarks}}
        & \multicolumn{1}{c|}{\textbf{Online video benchmark}}
        & \multirow{2}{*}{\textbf{Average}} \\
        \cmidrule(lr){3-5} \cmidrule(lr){6-6}
        & & VideoMME & LongVideoBench & EgoSchema
        & RVS-Ego & \\
        \midrule
        \midrule
        \multirow{2}{*}{InternVL3.5-8B}
        & Qwen3-VL-2B & 61.4 & \textbf{58.9} & 55.0 & 53.9 & \cellcolor{table6}{57.3} \\
        & InternVL3.5-2B & \textbf{61.5} & 58.5 & \textbf{55.6} & \textbf{54.1} & \cellcolor{table6}\textbf{57.4} \\
        \midrule
        \midrule
        \multirow{2}{*}{Qwen3-VL-8B}
        & Qwen3-VL-2B & \textbf{61.1} & 57.7 & 75.0 & 58.4 & \cellcolor{table6}{63.1} \\
        & InternVL3.5-2B & 60.6 & \textbf{58.0} & \textbf{76.2} & \textbf{58.8} & \cellcolor{table6}\textbf{63.4} \\
        \bottomrule
    \end{tabular}
    \end{adjustbox}
    \label{tab:qwen}
\end{table*}
\noindent \textbf{Effect of the reward model.}
In Tab.~\ref{tab:qwen}, we further assess the sensitivity of SVMemAgent to the reward model used during GRPO training by replacing InternVL3.5-2B with Qwen3-VL-2B. 
The policy trained with Qwen3-VL-2B achieves nearly identical performance, with average accuracy differences of only 0.1\% and 0.3\% when evaluated using InternVL3.5-8B and Qwen3-VL-8B as the downstream VideoLLMs, respectively. 
These results indicate that SVMemAgent learns a robust policy that is largely insensitive to the choice of reward model.

\begin{figure}[!t]
    \centering
    \includegraphics[width=\linewidth]{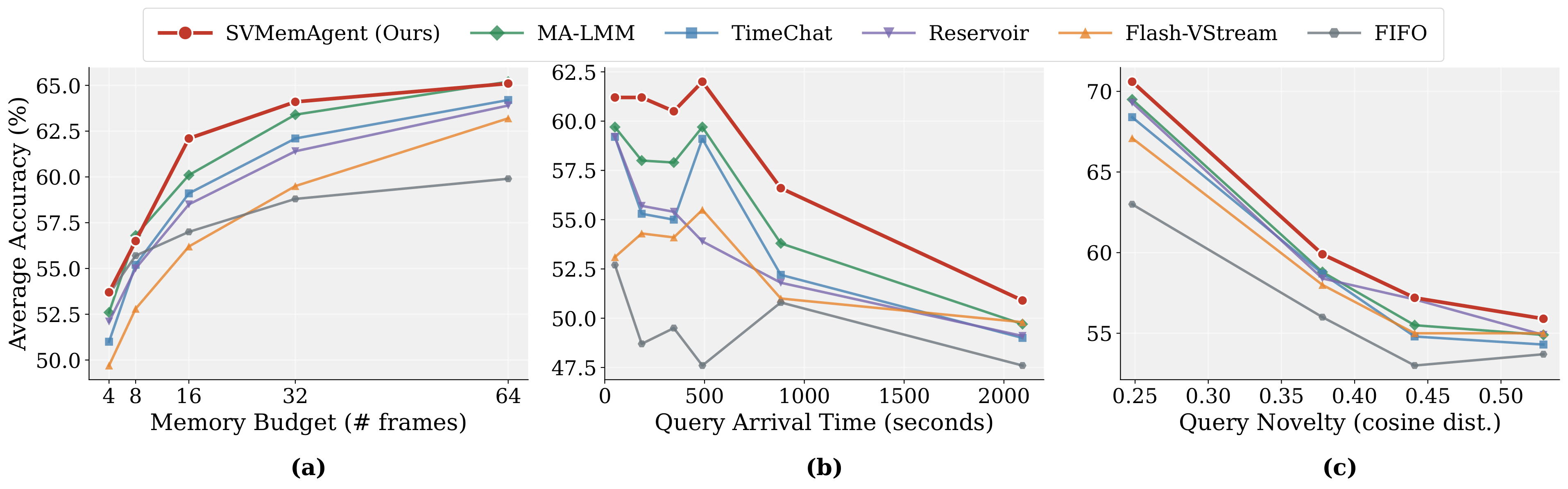}
    \caption{
        (a) Performance across different memory budgets, measured by the number of stored frames; (b) performance across different query arrival times; and (c) performance across varying levels of query novelty. 
        Query novelty is computed as the mean cosine distance between a test-query embedding and its 10 nearest training-query embeddings, obtained using Sentence-BERT~\cite{reimers2019sentence}. 
        Higher values indicate that the query is more out-of-distribution relative to the training queries.
    }
    \label{fig:three}
\end{figure}

\noindent \textbf{Memory budget scaling.}
Fig.~\ref{fig:three}a illustrates the effect of varying the inference-time memory budget from $K=4$ to $64$. 
SVMemAgent consistently outperforms competing online frame selection baselines across most memory budgets, despite being trained only under $K=16$.
These results suggest that the proposed policy learns scalable memory management behavior that generalizes across different memory capacities.

\subsection{Analysis}

To better understand why SVMemAgent performs effectively under online streaming constraints, we provide an in-depth analysis.

\noindent \textbf{Is SVMemAgent robust to varying query arrival times?}
Fig.~\ref{fig:three}b evaluates SVMemAgent across a wide range of query arrival times. 
Although the average query arrival time during training is 196 seconds, SVMemAgent consistently outperforms the baseline across all evaluated arrival times. 
This result indicates that the learned memory maintenance policy generalizes beyond the training-time arrival distribution, despite having no prior knowledge of when a user query will arrive, which highlights its applicability for real-world streaming applications.

\noindent \textbf{Does SVMemAgent generalize well to unseen queries?}
Since SVMemAgent performs frame selection without access to the query during streaming and is trained with task-driven rewards from only four queries per video, it is important to assess whether the learned policy remains robust under shifts in the query distribution.
To this end, we evaluate performance under shifts in the query distribution with respect to query novelty, defined as the mean cosine distance between each Sentence-BERT test-query embedding and its 10 nearest training-query embeddings~\cite{reimers2019sentence}.
As shown in Fig.~\ref{fig:three}c, SVMemAgent consistently outperforms all baselines across the full range of query novelty, indicating that the learned policy generalizes effectively even to queries that differ substantially from those encountered during training.

\begin{table*}[!t]
    \centering

    \begin{minipage}[t]{0.49\textwidth}
        \centering
        \captionof{table}{\textbf{Ablation studies on DAAD.}}
        \label{tab:daad}
        \vspace{3mm}

        \begin{adjustbox}{width=\linewidth}
        \begin{tabular}{l|cc|cc}
            \toprule
            \multirow{2}{*}{\textbf{Models}}
            & \multicolumn{2}{c|}{\textbf{InternVL3.5}}
            & \multicolumn{2}{c}{\textbf{Qwen3-VL}} \\
            & w/o DAAD
            & w/ DAAD
            & w/o DAAD
            & w/ DAAD \\
            \midrule
            \textbf{Average}
            & 56.3
            & \textbf{57.4}
            & 62.9
            & \textbf{63.4} \\
            \bottomrule
        \end{tabular}
        \end{adjustbox}
    \end{minipage}
    \hfill
    \begin{minipage}[t]{0.49\textwidth}
        \centering
        \captionof{table}{
            \textbf{Per-frame latency breakdown of SVMemAgent
            under the strict online setting.}
        }
        \label{tab:latency}
        \vspace{-2.5mm}

        \begin{adjustbox}{width=\linewidth}
        \begin{tabular}{l|ccc|c}
            \toprule
            & Image Enc.
            & Semantic
            & Temporal
            & \textbf{Total} \\
            \midrule
            \# of Params. (M)
            & 85.7
            & 86.8
            & 23.1
            & \textbf{195.6} \\
            Latency (ms/frame)
            & 7.0
            & 6.8
            & 2.2
            & \textbf{16.2 (61.7 FPS)} \\
            \bottomrule
        \end{tabular}
        \end{adjustbox}
    \end{minipage}
\end{table*}

\noindent \textbf{Which samples are discounted by DAAD?}
We observe that although each rollout contains a diverse set of frames, with $D_\text{Jac} = 0.9$, the resulting reward, defined as the negative loss, is already high and exhibits negligible variation across rollouts, with $\text{std}(R) = 0.000003$.
This indicates that the reward depends primarily on the query itself rather than on the frame selection policy. 
For example, a question asking why a person extends their arms while walking on a slackline can be answered from language bias alone, \ie, `to maintain balance'. 
Consequently, language-biased query-answer pairs introduce noisy supervision signals, which can destabilize policy optimization with the original advantages.
In contrast, DAAD successfully identifies such cases and assigns an advantage discounting score of $\alpha = 0$, reducing their advantages to $\hat{A} = 0$. 
This prevents noisy supervision from adversely affecting frame selection policy learning.
As in Tab.~\ref{tab:daad}, incorporating DAAD consistently improves performance over the variant without DAAD, demonstrating its effectiveness in stabilizing policy optimization.

\noindent \textbf{What policy does SVMemAgent learn?}
Given a video dominated by a man discussing his travel experiences, interspersed with a few short scenes, SVMemAgent implicitly learns to avoid redundant talking-head frames and instead preserve frames from less frequent scenes that are more likely to contain information relevant to downstream questions.
When prolonged talking scenes are already stored in memory, SVMemAgent repeatedly replaces the same memory slot with incoming frames depicting similar talking scenes.
It further discards such frames when the memory already contains sufficient talking-scene information.
As a result, the learned policy successfully retains a short segment in which the man shows a photograph, which is critical for answering the downstream question, ``Among the photos that the man is showing, which photo appears first?''. 
In contrast, offline uniform sampling repeatedly selects visually similar frames of the man speaking, resulting in a highly redundant memory state and an incorrect answer. 

Another intriguing behavior that emerges in SVMemAgent is its tendency to retain frames containing textual information.
This emergent behavior is learned through GRPO training with task-driven rewards across diverse question-answer pairs, suggesting that the agent recognizes textual content as decisive evidence for downstream VideoQA. 
For instance, given a video that depicts a person applying lipstick, and the downstream question asks for the product number of the first lipstick used. 
SVMemAgent identifies and retains the critical frame, which contains a product number, thereby enabling the VideoLLM to produce the correct answer.
In contrast, offline uniform sampling captures visually repetitive and less informative frames, ultimately leading to the incorrect answer. 
Overall, compared with offline uniform sampling, the proportion of text-containing frames stored in memory increases from 33.8\% to 40.8\%, suggesting that SVMemAgent implicitly recognizes textual content as a valuable source of information for future question answering and prioritizes such frames during memory maintenance.

\noindent \textbf{Efficiency analysis.}
During inference, the lightweight SVMemAgent with 195.6M parameters is invoked upon each frame arrival to update the memory using only the current memory state and the incoming frame. 
As in Tab.~\ref{tab:latency}, SVMemAgent achieves a memory maintenance speed of 61.7 FPS, corresponding to only 16.2 ms of additional latency per frame under the strict online setting.
\section{Conclusion}

We introduced \textbf{SVMem}, a memory formulation for strict online video understanding, where frames arrive sequentially, the total stream duration is unknown, and the query is unavailable during memory maintenance. 
Building on this formulation, we proposed \textbf{SVMemAgent}, which maintains a compact streaming memory through discard-and-replacement decisions. 
SVMemAgent is trained with GRPO using task-driven rewards, enabling it to learn a general prior for selecting informative frames without access to the query or future frames. 
Experiments on multiple video benchmarks demonstrate that SVMemAgent consistently outperforms online frame selection baselines and most offline methods, despite the latter assuming substantially stronger information access.


%
%
\bibliographystyle{splncs04}
\bibliography{main}

@String(CVPR  = {IEEE Conf. Comput. Vis. Pattern Recog.})

@String(ICCV  = {Int. Conf. Comput. Vis.})

@String(NeurIPS = {Adv. Neural Inform. Process. Syst.})

@String(ICLR  = {Int. Conf. Learn. Represent.})

@String(CVPR  = {CVPR})

@String(ICCV  = {ICCV})

@String(NeurIPS = {NeurIPS})

@String(ICLR  = {ICLR})

@article{bai2025qwen3,
  title={Qwen3-vl technical report},
  author={Bai, Shuai and Cai, Yuxuan and Chen, Ruizhe and Chen, Keqin and Chen, Xionghui and Cheng, Zesen and Deng, Lianghao and Ding, Wei and Gao, Chang and Ge, Chunjiang and others},
  journal={arXiv preprint arXiv:2511.21631},
  year={2025}
}

@inproceedings{ko2023large,
  title={Large language models are temporal and causal reasoners for video question answering},
  author={Ko, Dohwan and Lee, JiSoo and Kang, Woo-Young and Roh, Byungseok and Kim, Hyunwoo},
  booktitle={EMNLP},
  year={2023}
}

@article{li2024videochat,
  title={Videochat-flash: Hierarchical compression for long-context video modeling},
  author={Li, Xinhao and Wang, Yi and Yu, Jiashuo and Zeng, Xiangyu and Zhu, Yuhan and Huang, Haian and Gao, Jianfei and Li, Kunchang and He, Yinan and Wang, Chenting and others},
  journal={arXiv preprint arXiv:2501.00574},
  year={2024}
}

@article{zhang2024video,
  title={Video instruction tuning with synthetic data},
  author={Zhang, Yuanhan and Wu, Jinming and Li, Wei and Li, Bo and Ma, Zejun and Liu, Ziwei and Li, Chunyuan},
  journal={arXiv preprint arXiv:2410.02713},
  year={2024}
}

@article{zhang2025videollama,
  title={VideoLLaMA 3: Frontier Multimodal Foundation Models for Image and Video Understanding},
  author={Zhang, Boqiang and Li, Kehan and Cheng, Zesen and Hu, Zhiqiang and Yuan, Yuqian and Chen, Guanzheng and Leng, Sicong and Jiang, Yuming and Zhang, Hang and Li, Xin and others},
  journal={arXiv preprint arXiv:2501.13106},
  year={2025}
}

@inproceedings{li2025videochat,
  title={Videochat-r1: Enhancing spatio-temporal perception via reinforcement fine-tuning},
  author={Li, Xinhao and Yan, Ziang and Meng, Desen and Dong, Lu and Zeng, Xiangyu and He, Yinan and Wang, Yali and Qiao, Yu and Wang, Yi and Wang, Limin},
  booktitle={NeurIPS},
  year={2025}
}

@inproceedings{tang2025adaptive,
  title={Adaptive keyframe sampling for long video understanding},
  author={Tang, Xi and Qiu, Jihao and Xie, Lingxi and Tian, Yunjie and Jiao, Jianbin and Ye, Qixiang},
  booktitle={CVPR},
  year={2025}
}

@inproceedings{hu2025m,
  title={M-llm based video frame selection for efficient video understanding},
  author={Hu, Kai and Gao, Feng and Nie, Xiaohan and Zhou, Peng and Tran, Son and Neiman, Tal and Wang, Lingyun and Shah, Mubarak and Hamid, Raffay and Yin, Bing and others},
  booktitle={CVPR},
  year={2025}
}

@article{shao2024deepseekmath,
  title={Deepseekmath: Pushing the limits of mathematical reasoning in open language models},
  author={Shao, Zhihong and Wang, Peiyi and Zhu, Qihao and Xu, Runxin and Song, Junxiao and Bi, Xiao and Zhang, Haowei and Zhang, Mingchuan and Li, YK and Wu, Yang and others},
  journal={arXiv preprint arXiv:2402.03300},
  year={2024}
}

@article{wang2025internvl3,
  title={Internvl3. 5: Advancing open-source multimodal models in versatility, reasoning, and efficiency},
  author={Wang, Weiyun and Gao, Zhangwei and Gu, Lixin and Pu, Hengjun and Cui, Long and Wei, Xingguang and Liu, Zhaoyang and Jing, Linglin and Ye, Shenglong and Shao, Jie and others},
  journal={arXiv preprint arXiv:2508.18265},
  year={2025}
}

@inproceedings{chen2024videollm,
  title={Videollm-online: Online video large language model for streaming video},
  author={Chen, Joya and Lv, Zhaoyang and Wu, Shiwei and Lin, Kevin Qinghong and Song, Chenan and Gao, Difei and Liu, Jia-Wei and Gao, Ziteng and Mao, Dongxing and Shou, Mike Zheng},
  booktitle={CVPR},
  year={2024}
}

@inproceedings{di2025streaming,
  title={Streaming video question-answering with in-context video kv-cache retrieval},
  author={Di, Shangzhe and Yu, Zhelun and Zhang, Guanghao and Li, Haoyuan and Zhong, Tao and Cheng, Hao and Li, Bolin and He, Wanggui and Shu, Fangxun and Jiang, Hao},
  booktitle={ICLR},
  year={2025}
}

@article{yao2025timechat,
  title={TimeChat-Online: 80\% Visual Tokens are Naturally Redundant in Streaming Videos},
  author={Yao, Linli and Li, Yicheng and Wei, Yuancheng and Li, Lei and Ren, Shuhuai and Liu, Yuanxin and Ouyang, Kun and Wang, Lean and Li, Shicheng and Li, Sida and others},
  journal={arXiv preprint arXiv:2504.17343},
  year={2025}
}

@article{fu2025vispeak,
  title={ViSpeak: Visual Instruction Feedback in Streaming Videos},
  author={Fu, Shenghao and Yang, Qize and Li, Yuan-Ming and Peng, Yi-Xing and Lin, Kun-Yu and Wei, Xihan and Hu, Jian-Fang and Xie, Xiaohua and Zheng, Wei-Shi},
  journal={arXiv preprint arXiv:2503.12769},
  year={2025}
}

@article{qian2025dispider,
  title={Dispider: Enabling Video LLMs with Active Real-Time Interaction via Disentangled Perception, Decision, and Reaction},
  author={Qian, Rui and Ding, Shuangrui and Dong, Xiaoyi and Zhang, Pan and Zang, Yuhang and Cao, Yuhang and Lin, Dahua and Wang, Jiaqi},
  journal={arXiv preprint arXiv:2501.03218},
  year={2025}
}

@inproceedings{qian2024streaming,
  title={Streaming long video understanding with large language models},
  author={Qian, Rui and Dong, Xiaoyi and Zhang, Pan and Zang, Yuhang and Ding, Shuangrui and Lin, Dahua and Wang, Jiaqi},
  booktitle={NeurIPS},
  year={2024}
}

@inproceedings{li2025lion,
  title={Lion-fs: Fast \& slow video-language thinker as online video assistant},
  author={Li, Wei and Hu, Bing and Shao, Rui and Shen, Leyang and Nie, Liqiang},
  booktitle={CVPR},
  year={2025}
}

@article{liu2024streamchat,
  title={StreamChat: Chatting with Streaming Video},
  author={Liu, Jihao and Yu, Zhiding and Lan, Shiyi and Wang, Shihao and Fang, Rongyao and Kautz, Jan and Li, Hongsheng and Alvare, Jose M},
  journal={arXiv preprint arXiv:2412.08646},
  year={2024}
}

@inproceedings{xiong2025streaming,
  title={Streaming Video Understanding and Multi-round Interaction with Memory-enhanced Knowledge},
  author={Xiong, Haomiao and Yang, Zongxin and Yu, Jiazuo and Zhuge, Yunzhi and Zhang, Lu and Zhu, Jiawen and Lu, Huchuan},
  booktitle={ICLR},
  year={2025}
}

@inproceedings{huang2024online,
  title={Online Video Understanding: A Comprehensive Benchmark and Memory-Augmented Method},
  author={Huang, Zhenpeng and Li, Xinhao and Li, Jiaqi and Wang, Jing and Zeng, Xiangyu and Liang, Cheng and Wu, Tao and Chen, Xi and Li, Liang and Wang, Limin},
  booktitle={CVPR},
  year={2025}
}

@article{simeoni2025dinov3,
  title={Dinov3},
  author={Sim{\'e}oni, Oriane and Vo, Huy V and Seitzer, Maximilian and Baldassarre, Federico and Oquab, Maxime and Jose, Cijo and Khalidov, Vasil and Szafraniec, Marc and Yi, Seungeun and Ramamonjisoa, Micha{\"e}l and others},
  journal={arXiv preprint arXiv:2508.10104},
  year={2025}
}

@article{chen2024longvila,
  title={Longvila: Scaling long-context visual language models for long videos},
  author={Chen, Yukang and Xue, Fuzhao and Li, Dacheng and Hu, Qinghao and Zhu, Ligeng and Li, Xiuyu and Fang, Yunhao and Tang, Haotian and Yang, Shang and Liu, Zhijian and others},
  journal={arXiv preprint arXiv:2408.10188},
  year={2024}
}

@inproceedings{he2024ma,
  title={Ma-lmm: Memory-augmented large multimodal model for long-term video understanding},
  author={He, Bo and Li, Hengduo and Jang, Young Kyun and Jia, Menglin and Cao, Xuefei and Shah, Ashish and Shrivastava, Abhinav and Lim, Ser-Nam},
  booktitle={CVPR},
  year={2024}
}

@inproceedings{zhang2025flash,
  title={Flash-vstream: Efficient real-time understanding for long video streams},
  author={Zhang, Haoji and Wang, Yiqin and Tang, Yansong and Liu, Yong and Feng, Jiashi and Jin, Xiaojie},
  booktitle={ICCV},
  year={2025}
}

@inproceedings{feng2026video,
  title={Video-r1: Reinforcing video reasoning in mllms},
  author={Feng, Kaituo and Gong, Kaixiong and Li, Bohao and Guo, Zonghao and Wang, Yibing and Peng, Tianshuo and Wu, Junfei and Zhang, Xiaoying and Wang, Benyou and Yue, Xiangyu},
  booktitle={NeurIPS},
  year={2025}
}

@inproceedings{liu2025visual,
  title={Visual-rft: Visual reinforcement fine-tuning},
  author={Liu, Ziyu and Sun, Zeyi and Zang, Yuhang and Dong, Xiaoyi and Cao, Yuhang and Duan, Haodong and Lin, Dahua and Wang, Jiaqi},
  booktitle={ICCV},
  year={2025}
}

@inproceedings{feng2026onethinker,
  title={Onethinker: All-in-one reasoning model for image and video},
  author={Feng, Kaituo and Zhang, Manyuan and Li, Hongyu and Fan, Kaixuan and Chen, Shuang and Jiang, Yilei and Zheng, Dian and Sun, Peiwen and Zhang, Yiyuan and Sun, Haoze and others},
  booktitle={CVPR},
  year={2026}
}

@inproceedings{reimers2019sentence,
  title={Sentence-bert: Sentence embeddings using siamese bert-networks},
  author={Reimers, Nils and Gurevych, Iryna},
  booktitle={EMNLP},
  year={2019}
}
\end{document}